\documentclass[letterpaper,journal]{IEEEtran}
\usepackage{amsmath,amsfonts}
\usepackage[linesnumbered,ruled,vlined]{algorithm2e}
\SetKwInOut{Require}{Require}
\SetKw{KwReturn}{return}
\usepackage{array}
\usepackage[caption=false,font=normalsize,labelfont=sf,textfont=sf]{subfig}
\usepackage{textcomp}
\usepackage{stfloats}
\usepackage{url}
\usepackage{verbatim}
\usepackage{graphicx}
\usepackage{cite}
\usepackage{makecell}
\usepackage{multirow}
\usepackage{placeins}
\usepackage[table]{xcolor}  
\usepackage{makecell}       
\usepackage{pifont}         
\usepackage{xcolor}         
\usepackage{soul}

\begin{document}

\title{An Optimal Cascade Feature-Level Spatiotemporal Fusion Strategy for Anomaly Detection in CAN Bus}%

\author{
    Mohammad {Fatahi},~Danial {Sadrian Zadeh},~\IEEEmembership{Graduate Student Member,~IEEE,}~Benyamin {Ghojogh},~Behzad {Moshiri},~\IEEEmembership{Senior Member,~IEEE,}~Otman {Basir},~\IEEEmembership{Member,~IEEE}%
    \thanks{M. {Fatahi} and B. {Moshiri} are with the School of Electrical and Computer Engineering, College of Engineering, University of Tehran, Tehran, Tehran 1439957131, Iran.}%
    \thanks{D. {Sadrian Zadeh}, B. {Ghojogh}, B. {Moshiri}, and O. {Basir} are with the Department of Electrical and Computer Engineering, Faculty of Engineering, University of Waterloo, Waterloo, ON N2L 3G1, Canada.}%
    \thanks{Corresponding author: Behzad {Moshiri} (email: moshiri@ut.ac.ir)}%
}%



\maketitle%

\begin{abstract}
Intelligent transportation systems (ITS) play a pivotal role in modern infrastructure but face security risks due to the broadcast-based nature of the in-vehicle Controller Area Network (CAN) buses. While numerous machine learning models and strategies have been proposed to detect CAN anomalies, existing approaches lack robustness evaluations and fail to comprehensively detect attacks due to shifting their focus on a subset of dominant structures of anomalies. To overcome these limitations, the current study proposes a cascade feature-level spatiotemporal fusion framework that integrates the spatial features and temporal features through a two-parameter genetic algorithm (2P-GA)-optimized cascade architecture to cover all dominant structures of anomalies. Extensive paired t-test analysis confirms that the model achieves an AUC-ROC of 0.9987, demonstrating robust anomaly detection capabilities. The Spatial Module improves the precision by approximately 4\%, while the Temporal Module compensates for recall losses, ensuring high true positive rates. The proposed framework detects all attack types with 100\% accuracy on the CAR-HACKING dataset, outperforming state-of-the-art methods. This study provides a validated, robust solution for real-world CAN security challenges.%
\end{abstract}%


\section{Introduction}%
\label{sec:introduction}%
\subsection{Research Background and Motivations}%
\IEEEPARstart{T}he automotive industry, vital to global innovation and mobility, has shifted from mechanical components to Electronic Control Units (ECUs) over a decade ago to cut costs, boost efficiency, and enhance comfort \cite{2024_Rani}. This ongoing evolution continues to advance with ECUs growing increasingly complex \cite{HyunjaeKang2024}. ECUs rely on the widely recognized Controller Area Network (CAN) protocol for communication \cite{Agrawal2022}. The CAN bus message structure between ECUs includes a timestamp for recording time, CAN ID indicating message priority and identification, DLC for data length, and a data field for payload, thus enabling efficient communication \cite{Kang2021,Song2020}. The security weakness of CAN bus, due to its broadcast nature, priority-based CAN ID, and lack of built-in security mechanisms, makes it vulnerable to cyberattacks \cite{Agrawal2022,Alalwany2022}.%

These vulnerabilities pose significant risks to passenger and pedestrian safety, necessitating robust anomaly detection models to identify and mitigate cyber threats. One of the most widely adopted approaches for anomaly detection involves leveraging machine learning-based models \cite{2024_ALMahadin}. These models are trained on the patterns of CAN bus messages, which results in high generalizability. This high generalizability stems from the inherent consistency in the CAN bus message structure across different vehicle types with varying CAN architectures \cite{Nagarajan2023,2025_Fatahi}. Among the existing approaches, machine learning-based models are typically designed as black-box systems rather than being tailored to the underlying structure of the problem. These models often aim to improve accuracy by increasing architectural complexity. In response to these challenges, this study seeks to address the following research questions (RQ):%
\begin{itemize}
    \item \textbf{RQ1:} How can a machine learning model be designed based on the inherent structure of the problem, rather than treating it as a black-box and relying solely on added complexity to enhance performance?%
    \item \textbf{RQ2:} Can such a structure-aware model achieve superior performance metrics compared to conventional black-box approaches?%
\end{itemize}%

\subsection{Problem Statement}%
\begin{figure}[t]
     \centering%
     \includegraphics[width=\columnwidth]{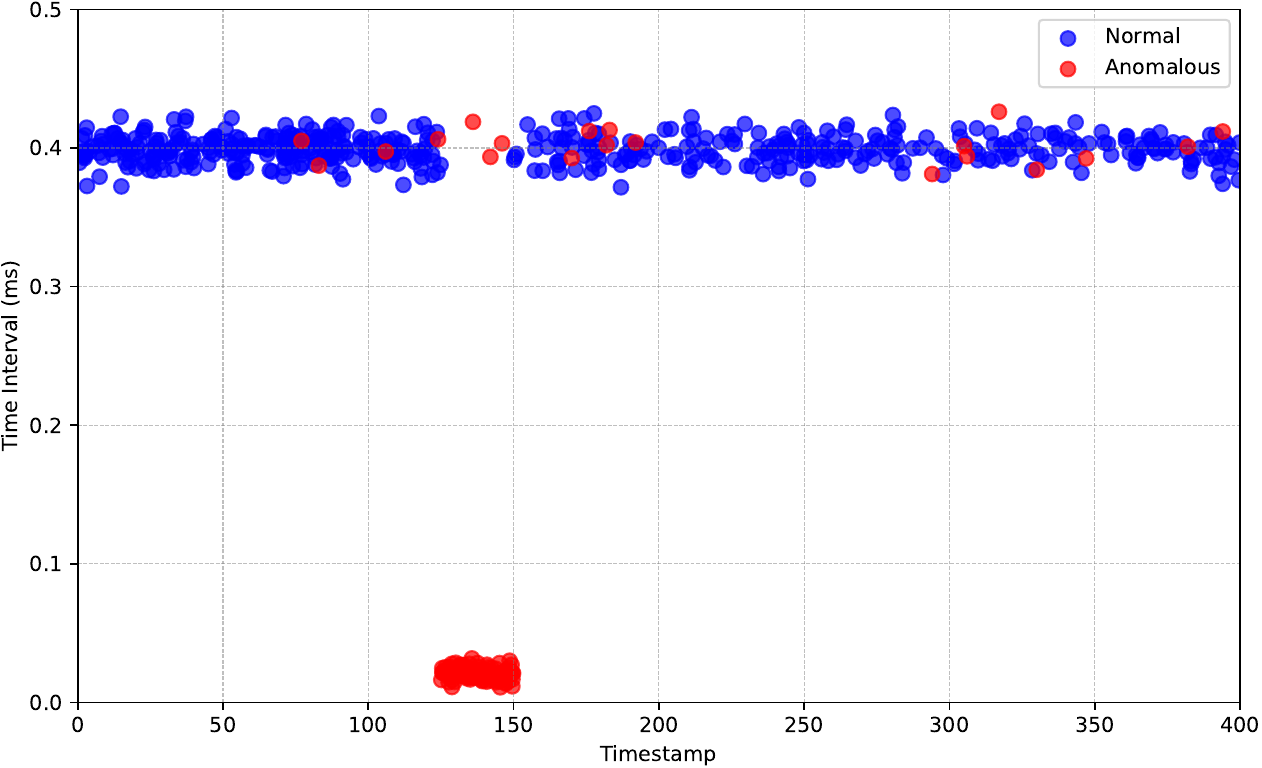}%
     \caption{Time interval analysis for CAN ID 0x490 during attack \cite{Kang2021}.}\label{TIMEINTER}%
\end{figure}

Attacks on the CAN bus exhibit temporal, spatial, and complex structures \cite{Kang2021}. Table~\ref{pattattack1} illustrates the spatial structure of attacks, in which the Data field changes during attacks on the CAN bus. Specifically, the data structure is altered by the attacker for Data1 and Data6 in Table~\ref{pattattack1}, potentially leading to severe consequences. Figure~\ref{TIMEINTER} depicts the temporal structure of attacks, where normal messages (in blue) have time intervals of approximately 0.4 milliseconds, whereas spoofed messages (in red) exhibit intervals of less than 0.04 milliseconds. The third structure, i.e., complex structure, involves, for instance, attackers injecting messages close to the defined inter-frame time or combining temporal and spatial structures for attacks. The current study proposes a structure-aware model that aims to develop a robust framework by considering these structures to counter such attacks.%

\begin{table}[t]
    \centering%
    \caption{Behavior Observed in CAN ID 0x490 During Spoofing Attack \cite{Kang2021}}\label{pattattack1}%
    \resizebox{\columnwidth}{!}{%
    \begin{tabular}{|c|c|c|c|c|c|c|c|c|c|c|}
        \hline
        \textbf{Timestamp} & \textbf{CAN ID} & \textbf{DLC} & \textbf{Data1} & \textbf{Data2} & \textbf{Data3} & \textbf{Data4} & \textbf{Data5} & \textbf{Data6} & \textbf{Data7} & \textbf{Data8}\\%
        \hline
        120.78377 & 490 & 8 & 00 & 00 & 08 & 21 & 00 & 00 & 3C & 7C\\%
        120.83318 & 490 & 8 & 00 & 00 & 08 & 21 & 00 & 10 & 3C & C8\\%
        120.88338 & 490 & 8 & 00 & 00 & 08 & 21 & 00 & 20 & 3C & 09\\%
        \rowcolor[HTML]{FFCCCC} 
        120.90583 & 490 & 8 & 03 & 00 & 08 & 21 & 00 & 00 & 3C & 7C\\%
        120.93318 & 490 & 8 & 00 & 00 & 08 & 21 & 00 & 30 & 3C & BD\\%
        120.98348 & 490 & 8 & 00 & 00 & 08 & 21 & 00 & 40 & 3C & 96\\%
        120.03387 & 490 & 8 & 00 & 00 & 08 & 21 & 00 & 50 & 3C & 22\\%
        \hline
    \end{tabular}%
    }%
\end{table}%

\subsection{Research Contributions}%
In this regard, the key contributions of this manuscript are outlined as follows:%
\begin{itemize}
    \item \textbf{Cascade feature-level spatiotemporal fusion strategy:} A novel cascade feature-level fusion approach is proposed, hierarchically integrating raw, spatial, and temporal features to enhance detection accuracy. This method addresses the limitations of single-dimensional and conventional black-box models by capturing complex attack patterns effectively.%
    \item \textbf{Structure-aware and explainable modeling:} The proposed model considers the attack structures and provides explainable insights into how modeling each component of the attack pattern contributes to the overall performance improvement. This transparency aids in understanding the model’s decision-making process and enhances its reliability in safety-critical systems.%
    \item \textbf{Comprehensive analysis and statistical validation:} This work conducts an in-depth evaluation of spatial feature synthesis to improve anomaly detection in CAN messages, leveraging recurrent neural networks (RNNs) for future message prediction. By analyzing the structure of the Data field, the model enhances the detection of safety-critical manipulations. Additionally, the study provides statistically validated results, including paired \(t\)-tests, classifier comparisons, AUC-ROC, inference time, and a comprehensive comparison with state-of-the-art and feature engineering-based approaches.%
\end{itemize}%

The remainder of this manuscript is structured as follows: Section~\ref{sec:s02-related-work} reviews the related work. Section~\ref{sec:s03-methodology} elaborates on the proposed framework and its stages, while Section~\ref{sec:s04-analysis-of-simulation-results} presents the results, followed by their analysis. Finally, Section~\ref{sec:s05-conclusion-and-future-works} concludes the manuscript and proposes directions for future research.

\section{Related Work}%
\label{sec:s02-related-work}%
Anomaly detection in intelligent transportation systems (ITS) involves a diverse set of techniques. These techniques have been widely examined to address the challenges of identifying anomalies in the CAN bus. Table~\ref{litcom} summarizes the key findings and limitations of some of the state-of-the-art studies in this area of research. As shown in Table~\ref{litcom}, all studies conducted in this area of research lack statistical testing.%

\begin{table}[t]
    \centering
    \caption{Summary of Related Work}\label{litcom}
    \begin{tabular}{|c|c|c|c|c|}
        \hline
        \textbf{Reference} & \textbf{Year} & \textbf{Dataset} & \textbf{\makecell{Input\\Features}} & \textbf{\makecell{Statistical\\Test}}\\%
        \hline
        \cite{Ashraf2021} & {2021} & {\cite{Song2020}} & {\makecell{CAN ID\\Timestamp\\DLC}} & {\textcolor{red}{\ding{55}}}\\%
        \hline
        \cite{Song2021} & {2021} & {\cite{Song2020}} & {CAN ID} & {\textcolor{red}{\ding{55}}}\\%
        \hline
        \cite{Alalwany2022} & {2022} & {\cite{Kang2021}} & {All} & {\textcolor{red}{\ding{55}}}\\%
        \hline
        \cite{Agrawal2022} & {2022} & {\cite{Song2020}} & {\makecell{CAN ID\\Data}} & {\textcolor{red}{\ding{55}}}\\%
        \hline
        \cite{Driss2022} & {2022} & {\cite{Kang2021}} & {All} & {\textcolor{red}{\ding{55}}}\\%
        \hline
        \cite{Aksu2022} & {2022} & {\cite{Kang2021}} & {All} & {\textcolor{red}{\ding{55}}}\\%
        \hline
        \cite{wei2023novel} & {2023} & {\cite{lee2017otids}} & {Data} & {\textcolor{red}{\ding{55}}}\\%
        \hline
        \cite{Mansourian2023} & {2023} & {\cite{Song2020}} & {Data} & {\textcolor{red}{\ding{55}}}\\%
        \hline
        \cite{Raj2023} & {2023} & {\cite{Kang2021}} & {All} & {\textcolor{red}{\ding{55}}}\\%
        \hline
        \cite{WeifengGong2024} & {2024} & {\cite{Song2020}} & {All} & {\textcolor{red}{\ding{55}}}\\%
        \hline
        \cite{kristianto2024sustainable} & {2024} & {\cite{Song2020}} & {All} & {\textcolor{red}{\ding{55}}}\\%
        \hline
    \end{tabular}%
\end{table}%

Prior studies adopted different strategies for dealing with attacks. For example, the work of \cite{wei2023novel} primarily focused on the Data field while disregarding the CAN ID and Timestamp features. Their approach utilized an attention-based network to extract relationships in the Data field. However, its most significant drawback lies in neglecting the temporal characteristics of the information. The work of \cite{WeifengGong2024} developed a model by leveraging an end-to-end approach where multi-level features were extracted using an attention-based network, factorization machine (FM), and Cross Network. Despite the increased model complexity, the performance in terms of accuracy and F1-score fell short of expectations on a dataset where near-perfect performance is typically achievable. This indicates that the high complexity and diverse components of the network failed to collaborate effectively to deliver optimal results. In \cite{Mansourian2023}, the focus of the authors was on spatial information, relying solely on the Data field, while attempting to predict anomalies in a supervised manner using a long short-term memory (LSTM) network and a Gaussian Naive Bayes classification technique. This study also overlooked the importance of temporal information and developed a model relying solely on spatial information, which significantly weakened the performance of the model. In \cite{kristianto2024sustainable}, similar to the work of \cite{Mansourian2023}, there was an emphasis on dealing with attacks through spatial differences via the employment of RNNs followed by the ReLU activation function. However, the issue lies in the fact that these spatial differences alone are insufficient for dealing with attacks, as they neglect the temporally dominant pattern of attacks. In addition, various experiments have revealed that employing ReLU in RNNs can lead to unstable training due to exploding gradients.%

In \cite{Alalwany2022}, the synthetic minority over-sampling technique (SMOTE) \cite{Chawla2002} was incorrectly applied to the entire dataset, which changed the natural distribution of the test set. This is because of the fact that SMOTE should be applied to the training set, not the test set. Furthermore, multi-classifier fusion was used in the framework to combine various classifiers for classification. In \cite{Agrawal2022}, the authors proposed a model consisting of two LSTM layers, each with 256 units, one one-dimensional convolution neural network (1D-CNN) layer with 100 filters, and a fully connected layer with nine neurons at the end. Their proposed model processed sequences of length 100 and made predictions by comparing them with a threshold. First, such a model is computationally demanding and requires very powerful hardware, which may not be cost-effective in industrial environments. Moreover, the results of the proposed model were analyzed only on a single dataset, which highlights the need for further evaluation. In \cite{Ashraf2021}, the proposed methodology solely focused on the temporal aspect of attacks on the CAN bus, neglecting the spatial aspect, which constitutes one of the two dominant attack patterns in this domain. Additionally, the accuracy metric was not reported, and the results were not presented for different parts of the dataset. In \cite{Driss2022}, the authors utilized a federated learning (FL) framework, where data was distributed across different nodes, and each node trained a gated recurrent unit (GRU) network. The weights of the local models were then aggregated on a central server to create a global model using an aggregation technique and an ensemble unit to improve accuracy. Similar to \cite{Agrawal2022}, the approach of \cite{Driss2022} is computationally demanding, which may pose a challenge in industrial applications. The framework in \cite{Aksu2022} includes feature selection and the selection of efficient classifiers. While such a simple model is computationally efficient, it yielded mediocre results. In \cite{Raj2023}, after performing oversampling, various classifiers were applied, which were computationally effective but exhibited poor performance. The proposed model in \cite{Song2021} relied solely on CAN ID, thus overlooking the Data field, which significantly weakens the model as it neglects the dominant spatial pattern. The proposed model consisted of two parts: a generator and a detector. While the former employed an LSTM network to produce pseudo-normal noisy data from CAN ID sequences, trying to reconstruct the characteristics of normal traffic, the latter leveraged a ResNet \cite{2015_He_ResNet} for classification.%

Overall, there are two key factors regarding the background of the research that need to be highlighted. These factors are as follows:%
\begin{enumerate}
    \item \textbf{Lack of Statistical Tests:} Most studies lack statistical tests, whereas providing such information significantly enhances the robustness and reliability of the research.%
    \item \textbf{Focus on a Single Aspect of the Dataset:} A significant number of studies either focused on the spatial aspect \cite{wei2023novel,Mansourian2023} or the temporal aspect \cite{Ashraf2021,Song2021}. Others aimed to develop models using all features in the dataset. However, these models failed to be robust to all attack structures. The optimal solution to overcome this challenge is to develop a model tailored to these structures.%
\end{enumerate}

\section{Methodology}%
\begin{figure*}[t]
    \centering
    \includegraphics[width=1.00\textwidth]{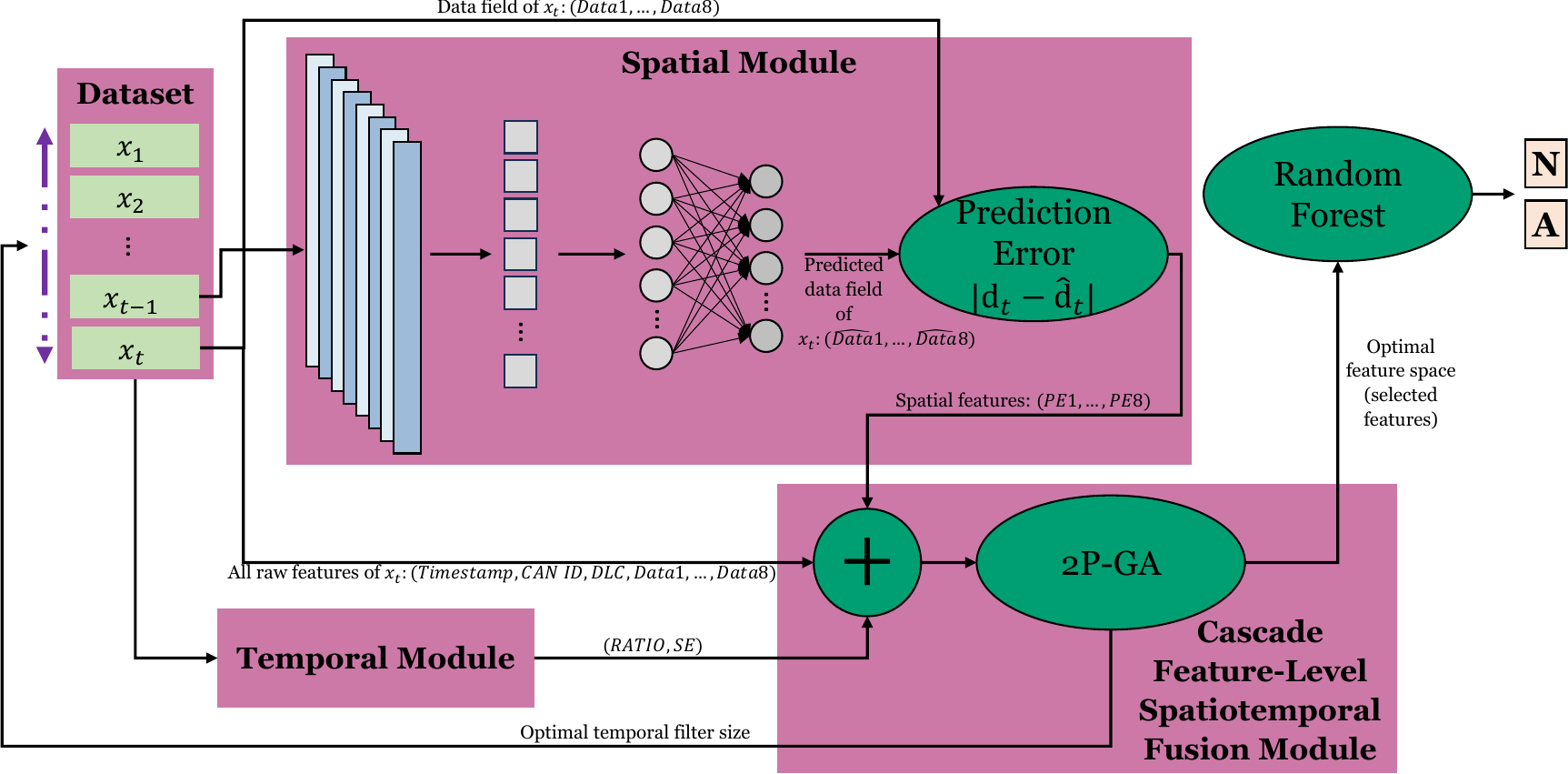}%
    \caption{Overall architecture of the proposed framework.}%
    \label{pippo}%
\end{figure*}

\label{sec:s03-methodology}%
This section provides a detailed explanation of the proposed methodology. Figure~\ref{pippo} depicts the proposed methodology, which consists of three main modules: (i) \textit{Spatial Module}, (ii) \textit{Temporal Module}, and (iii) \textit{Cascade Feature-Level Spatiotemporal Fusion Module}.%

\subsection{Spatial Module}%
\label{S&C}%
The main purpose of the Spatial Module stems from the spatially dominant structure of attacks carried out in the CAN bus mentioned in Section~\ref{sec:introduction}. In the spatial structure of attacks, the attacker's manipulations are applied to the values within the Data field. Accordingly, the objective of the current study is to enhance the ability of the model to accurately detect such structural patterns. Figure~\ref{pippo} illustrates that the Spatial Module consists of a network based on the 1D-CNN that is first trained on the Attack-Free/Normal dataset introduced in \cite{Song2020}. Notably, this part of the dataset is exclusively employed for training of this model and is not used in further analysis. Subsequently, this model is employed as a pre-trained model in the pipeline to predict the current message ($x_t$) solely based on the previous message ($x_{t-1}$). This process involves predicting eight values (Data1 to Data8), followed by calculating the absolute difference between each predicted value and its corresponding actual value received from the CAN bus. Finally, these differential values are utilized as the new features. Additionally, for the sake of comparison, the performance of the model is evaluated by replacing 1D-CNN with RNN \cite{Marsetic2024}, LSTM network \cite{1997_Hochreiter_LSTM}, GRU network \cite{RahulDey2017}, and xLSTM network \cite{Beck2024} for extracting spatial features. The details of the aforementioned networks used in the current study for extracting spatial features are as follows:%
\begin{itemize}
    \item \textbf{RNN} consists of a single-layer RNN with 64 units, a \texttt{tanh} activation function, and a recurrent dropout of 0.25. This is followed by a dense layer with eight neurons.%
    \item \textbf{LSTM network} is a specialized type of RNN designed to address the limitations of RNNs, particularly in capturing long-term dependencies \cite{Fatahi2023,Mansourian2023}. In this study, a single-layer LSTM \cite{1997_Hochreiter_LSTM} with the same configuration as the RNN is employed.%
    \item \textbf{GRU network} is a simplified version of LSTM that requires fewer parameters while still addressing vanishing gradients \cite{RahulDey2017}. Similar to the LSTM configuration, a single-layer GRU is employed.%
    \item \textbf{1D-CNN} is effective for sequential data by applying convolutional filters to extract local patterns \cite{Arsalan2024}. The configuration is similar to other networks, but the activation function is set to \texttt{ReLU}.%
    \item \textbf{xLSTM network}  is an enhanced LSTM architecture addressing its limitations through exponential gating and advanced memory structures. It includes two models: (i) \textit{sLSTM},  using scalar memory with exponential gating and memory mixing for improved information flow, and (ii) \textit{mLSTM}, which uses a matrix memory and a covariance update rule for fully parallelizable operations \cite{Beck2024}. This model, in the current study, is leveraged with an embedding size of 64 and four heads. It is supported by a Conv1D kernel of size four for 1D-CNN and a QKV projection block size of four governing the query, key, and value projections in the mLSTM/sLSTM.%
\end{itemize}%

\subsection{Temporal Module}%
\label{T&C}%
This module stems from the temporally dominant structure of attacks carried out in the CAN bus mentioned in Section~\ref{sec:introduction}. To enhance the capability of the model for dealing with the alteration in the share of each unique CAN ID within specific time intervals, similar to \cite{2025_Fatahi}, the Shannon entropy (SE) is leveraged as a part of the feature extraction process, leading to employment of two novel features proposed in \cite{2025_Fatahi}. These features are as follows (for detailed information on these features, the reader can refer to \cite{2025_Fatahi}): (i) \textit{RATIO}, which is the proportion of various CAN IDs within each time interval relative to the total number of messages during that interval, and (ii) \textit{SE}, which is the entropy-based feature for each CAN ID in a data sequence and is expressed as:%
\begin{equation}\label{eqn:sh2}
    \text{SE}(\text{ID}_{j}) = \frac{P(\text{ID}_{j}) \cdot \log_{2} P(\text{ID}_{j})}{\sum_{i=1}^{N} P(\text{ID}_{i}) \cdot \log_{2} P(\text{ID}_{i})}\text{,}%
\end{equation}%
where $\text{SE}(\text{ID}_{j})$ denotes the entropy-based feature for a specific CAN ID ($\text{ID}_{j}$) in the sequence, $P(\text{ID}_{j})$ stands for the proportion of messages corresponding to $\text{ID}_{j}$ in the total number of messages in the sequence, and $N$ represents the total number of unique CAN IDs in the sequence.%

\subsection{Cascade Feature-Level Spatiotemporal Fusion Module}%
\label{OCFLF}%
The primary purpose of fusion methods is to integrate various types of data to achieve higher performance. These methods can be divided into three main levels \cite{Martin2009}: (i) \textit{signal level}, for combining raw data before processing; (ii) \textit{feature level}, for merging extracted features from data to create a unified representation; and (iii) \textit{decision level}, for aggregating final results from multiple models or algorithms to make the ultimate decision.%

Feature-level fusion aims to enhance the ability of the model to extract deeper information by directly combining various features obtained from data or different layers of a network. An advanced technique within this domain is cascade feature-level fusion, which hierarchically integrates the features. In this approach, features are combined at different levels of abstraction, where higher-level features and lower-level ones are integrated to capture complex patterns \cite{ChuanxuWang2023}.%

In the current study, as shown in Fig.~\ref{pippo}, a cascade feature-level fusion approach is proposed to enhance the capability of the model for dealing with temporal, spatial, and complex structures of the attacks. In this regard, spatial (high-level) and temporal (low-level) features are concatenated with the raw features (low-level) from the dataset to create a cascade feature-level fusion framework. However, two questions arise here. First, what filter size should be used to extract temporal features? Second, what set of features is more effective? The point is that the performance of temporal features depends on the filter size; therefore, finding an optimal temporal filter size and feature selection must be performed simultaneously. Unfortunately, feature selection algorithms such as Recursive Feature Elimination (RFE) are unable to perform simultaneous extraction and selection. Therefore, in the current study, a two-parameter genetic algorithm (2P-GA) is proposed to simultaneously select the appropriate filter size for temporal features as well as appropriate features from the feature space (feature selection). This 2P-GA is leveraged with a population size of 25, a crossover rate of 0.9, a mutation rate of 0.1, and a generation number of five. These values for mutation and crossover represent probabilistic values for controlling the genetic operations. Also, the fitness function of the 2P-GA is as follows:%
\begin{equation}\label{eqn:fitness}
    \text{Fitness} = \text{F1-score} - (0.001 \times \text{num\_selected\_features})\text{.}%
\end{equation}%

The reason for opting for the F1-score as the criterion is that the datasets in this area of research are imbalanced, and the number of normal samples is several times greater than that of anomalous samples. In this case, the F1-score is a more reliable metric for evaluating the performance of the model in detecting anomalies. Additionally, incorporating a penalty on the number of selected features aims to achieve a compact feature set for enhanced efficiency. Algorithm~\ref{alg:optimal_subspace} presents the strategy for optimal cascade feature-level fusion. In this algorithm, a Decision Tree (DT) was used to examine the performance of each proposed feature subset and calculate the fitness value. It has been shown that DT has a very short training and inference time in this task and provides promising performance in terms of accuracy \cite{Aksu2022,Alalwany2022,2025_Fatahi}. Therefore, due to hardware limitations, the long training time of other classifiers, such as Random Forest (RF), and the need to repeat the training of the model for each feature subset, the present study employs a DT within the 2P-GA. However, after finding the optimal filter size and feature subset by the 2P-GA, RF is leveraged for the final prediction, as shown in Fig.~\ref{pippo}. The reason for this choice is that RF has better performance than other methods in terms of inference time, complexity, and accuracy. This statement is supported by the findings provided in the subsequent section.%

\begin{algorithm}[!t]
    \caption{Optimal Cascade Feature-Level Fusion}%
    \label{alg:optimal_subspace}%
    \small%
    \Require{Raw Features, Spatial Features, Temporal Features}%
    \SetAlgoLined%
    Preprocess raw features\;%
    Extract spatial features\;%
    Initialize population with random filter sizes and binary feature selectors\;%
    \While{maximum generations not reached}{%
        \For{each individual in population}{%
            Decode filter size from binary values\;%
            Apply filter to extract temporal features\;%
            Select features based on individual's binary mask\;%
            Train DT classifier\;%
            Compute F1-score on validation set\;%
            Compute fitness using \eqref{eqn:fitness}\;%
        }%
        Select 80\% of non-elite individuals using tournament selection and 20\% randomly to maintain diversity\;%
        Preserve top 20\% of population (elite individuals) for direct transfer to next generation\;%
        Apply crossover with probability 0.9 to non-elite selected individuals\;%
        Apply mutation with probability 0.1 to non-elite selected individuals\;%
        \If{converged}{%
            \KwRet{Best individual and break the loop}\;%
        }%
    }%
    \KwReturn{Best individual (optimal filter size and selected features)}\;%
\end{algorithm}

\section{Results and Discussion}%
\label{sec:s04-analysis-of-simulation-results}%
\subsection{Datasets}%
\label{data}%
As highlighted in Section~\ref{sec:s02-related-work}, the limited availability of datasets poses a significant challenge. Irreparable damages and potential hazards have resulted in a scarcity of real-world tests in research works in the field of cybersecurity for autonomous vehicles \cite{Nagarajan2023}. The Hacking and Countermeasure Research Lab (HCRL) \cite{HCRLwebsite2024} has provided valuable real-world datasets for advancing research in this field. The current study exploits two datasets from this laboratory to demonstrate and validate the proposed methodology. The first dataset, introduced in \cite{Kang2021}, is employed to illustrate the proposed methodology step-by-step in Section~\ref{sec:s04-analysis-of-simulation-results}, as it is the most recent and fully labeled dataset available on this subject. For the final evaluation in Subsection~\ref{CAWOSAM}, the dataset presented in \cite{Song2020} is utilized. Table~\ref{tab:combined_distribution} presents the distribution of anomalous messages in the datasets based on the type of attack and the number of normal messages transmitted.%

\begin{table}[t]
    \centering
    \caption{Distribution of Messages in Datasets}
    \label{tab:combined_distribution}
    \resizebox{\columnwidth}{!}{%
    \begin{tabular}{|l|l|l|l|l|}
        \hline
        \textbf{Dataset} & \textbf{Dataset Partition} & \textbf{Normal} & \textbf{Injected} & \textbf{Total}\\%
        \hline
        First Dataset \cite{Kang2021} & Normal \& Anomalous & 3,372,743 & 299,408 & 3,672,151\\%
        \hline
        \multirow{6}{*}{Second Dataset \cite{Song2020}} & DoS Attack & 3,078,250 & 587,521 & 3,665,771\\%
                                                        & Fuzzy Attack & 3,347,013 & 491,847 & 3,838,860\\%
                                                        & Spoofing Drive Gear & 3,845,890 & 597,252 & 4,443,142\\%
                                                        & Spoofing RPM Gauge & 3,966,805 & 654,897 & 4,621,702\\%
                                                        & GIDS (Attack-Free) & 988,872 & 0 & 988,872\\%
                                                        & All & 15,226,830 & 2,331,517 & 17,558,347\\%
        \hline
    \end{tabular}%
    }%
\end{table}%

\subsection{Data Preprocessing}%
\label{dp_ana}%
The data preprocessing stage consists of two steps. The first step replaces the \texttt{NaN} values with 0 and converts the hexadecimal values to decimal values. The second step divides the datasets into training (70\%), validation (15\%), and test (15\%) sets before normalizing all features, which reduces the impact of features with varying scales.%

\subsection{Simulation Results}%
\label{S&C_ana}%
Analysis of the results of the proposed framework on the dataset presented in \cite{Kang2021} is provided here. Table~\ref{tab:MAEPARAM} illustrates the performance of various models in predicting the current message based on the previous message, trained on the Attack-Free/Normal dataset presented in \cite{Song2020}. All models exhibit comparable performance in terms of mean absolute error (MAE), with the GRU-based model slightly outperforming others. However, when considering the number of parameters and inference time (measured in milliseconds per sample), the 1D-CNN-based model demonstrates superior efficiency. Consequently, selecting the 1D-CNN is justified due to its acceptable MAE index, coupled with lower computational complexity and faster inference time.%

\begin{table}[t]
    \caption{Performance Comparison for Each Network Tested as Spatial Data Prediction Model on CPU Based on \cite{Song2020}}
    \label{tab:MAEPARAM}
    \centering
    \begin{tabular}{|l|l|l|l|}
        \hline
        \textbf{Model} & \textbf{MAE} & \textbf{Parameters} & \textbf{Time}\\%
        \hline
        RNN & 0.4449 & 5,384 & 0.0479\\%
        \hline
        LSTM & 0.4362 & 19,976 & 0.0663\\%
        \hline
        GRU & \textbf{0.4353} & 15,304 & 0.0545\\%
        \hline
        xLSTM & 0.4359 & 32,208 & 0.0782\\%
        \hline
        1D-CNN & 0.4586 & \textbf{2,696} & \textbf{0.0433}\\%
        \hline
    \end{tabular}%
\end{table}%

Table~\ref{HF} shows that applying 2P-GA yields a filter size of 9332 with a fitness of 0.9488. Note that the features \textit{PE4} and \textit{PE6} are from the set of features generated by the Spatial Module. Table~\ref{tab:fusion_comparison} presents an analysis of the proposed framework. Initially, the performance of the RF classifier is evaluated on all raw features of the dataset. Subsequently, the performance of this classifier is assessed on a combination of all raw features and all spatial features, termed ``spatial fusion,'' yielding a 0.14\% increase in accuracy compared with raw features. More notably, the spatial fusion yields a significant improvement of approximately 4\% in precision. However, recall experiences a decline of nearly 2\%. In the next step, the temporal fusion is investigated, where all raw features are combined with the two proposed temporal features (i.e., \textit{SE} and \textit{RATIO}). To extract temporal features, a filter size of 7500 is applied, as proposed in \cite{Ashraf2021}. This temporal combination leads to approximately 1\% improvement in precision and a marginal 0.05\% enhancement in recall. The performance of various classifiers under the spatiotemporal fusion is compared in the last part of Table~\ref{tab:fusion_comparison}. The proposed framework with RF classifier achieves the highest accuracy among all classifiers, reaching 0.9952, which reflects a 0.3\% improvement over raw features. The precision of this model is approximately 4\% higher than that of the model with raw features, closely aligning with the performance of spatial fusion and spatiotemporal fusion using LightGBM and XGBoost. Recall, on the other hand, shows a 0.2\% increase, indicating that combining temporal and spatial features compensates for the recall reduction observed in spatial fusion while maintaining comparable precision. Therefore, the Spatial Module significantly enhances precision by reducing false positives, while the temporal module effectively compensates for the lower recall of the Spatial Module by improving the detection of true positives.%

\begin{table}[t]
    \centering
    \caption{Optimal Feature Space Based on \cite{Kang2021}}%
    \label{HF}%
    \resizebox{\columnwidth}{!}{%
    \begin{tabular}{|l|l|l|}
        \hline
        \textbf{Filter Size} & \textbf{Optimal Features} & \textbf{Fitness}\\%
        \hline
        9332 & \makecell[l]{Timestamp, CAN ID, Data3, Data4, Data5, Data6,\\Data7, Data8, SE, RATIO, PE4, PE6} & 0.9488\\%
        \hline
    \end{tabular}%
    }%
\end{table}%

\begin{table}[t]
    \centering
    \caption{Performance Comparison for Different Methods Trained on \cite{Kang2021}}
    \label{tab:fusion_comparison}
    \resizebox{\columnwidth}{!}{%
    \begin{tabular}{|l|l|l|l|l|l|l|}
        \hline
        \textbf{Features} & \textbf{Classifier} & \textbf{Accuracy} & \textbf{Precision} & \textbf{Recall} & \textbf{F1-score} & \textbf{Time}\\%
        \hline
        Raw & RF & 0.9923 & 0.9583 & 0.9466 & 0.9524 & \textbf{0.01365}\\%
        \hline
        Spatial Fusion& RF & 0.9937 & 0.9948 & 0.9274 & 0.9599 & 0.05938\\%
        \hline
        Temporal Fusion& RF & 0.9927 & 0.9633 & 0.9471 & 0.9551 & 0.01481\\%
        \hline
        \multirow{7}{*}{Spatiotemporal Fusion} & DT & 0.9934 & 0.9604 & \textbf{0.9588} & 0.9596 & 0.04342\\%
                                               & RF & \textbf{0.9952} & 0.9933 & 0.9484 & \textbf{0.9703} & 0.05762\\%
                                               & XGBoost & 0.9932 & 0.9969 & 0.9198 & 0.9568 & 0.04381\\%
                                               & LightGBM & 0.9916 & \textbf{0.9972} & 0.9001 & 0.9462 & 0.04475\\%
                                               & LR& 0.9561 & 0.9918 & 0.4686 & 0.6365 & 0.04331\\%
                                               & Naïve Bayes & 0.9583 & 0.9419 & 0.5241 & 0.6734 & 0.04338\\%
                                               & MLP & 0.9848 & 0.9862 & 0.8268 & 0.8995 & 0.04354\\%
        \hline
    \end{tabular}%
    }%
\end{table}%

\subsection{Robustness and Reliability}%
The current study employs the 5x2 cross-validated (5x2cv) paired \(t\)-test method \cite{1998_Dietterich} to examine the reliability of the proposed framework. This test is a robust statistical technique for comparing two machine learning models. It performs five rounds of two-fold cross-validation, where in each round the dataset is split in half, alternating between training and testing. This yields ten performance scores, with the paired \(t\)-test applied to their differences. The ultimate goal of applying this test is to determine the significance of the proposed framework based on RF against all raw features with RF. The test yields a \(t\)-statistic of 22.83 and a p-value of $1.17 \times 10^{-6}$, indicating that the proposed methodology is statistically significant.%

\subsection{Comparative Analysis With Other Feature Engineering Methods}%
Comparing the proposed framework with other feature engineering methods is essential for evaluating the effectiveness of different approaches in managing the feature space to achieve optimal performance. To this end, several networks---including Deep Neural Network (DNN), LSTM, attention-based network, Cross Network, and FM---were trained on the dataset in \cite{Kang2021}. As displayed in Table~\ref{featureinvestigate}, the proposed framework achieves the highest performance across all evaluation metrics, except for precision. The DNN, Wide and Deep Network, LSTM, and Cross Network exhibit very high precision; however, they demonstrate significantly lower recall compared with the proposed framework in the current study. In fact, such networks tend to make less effort in announcing an instance as an anomaly, which results in higher precision but poor performance in terms of recall. In contrast, the FM performs the worst, particularly in recall and F1-score, indicating its limited ability to capture the underlying feature interactions. While LSTM slightly improves recall, it comes at the cost of increased inference time. Overall, the proposed framework not only outperforms other approaches in effectiveness but also maintains a low computational footprint, making it suitable for practical deployment.%

\begin{table}[!t]
    \centering
    \caption{Performance Comparison for Feature Engineering Methods Trained on \cite{Kang2021}}%
    \label{featureinvestigate}%
    \resizebox{\columnwidth}{!}{%
    \begin{tabular}{|l|l|l|l|l|l|l|}
        \hline
        \textbf{Method} & \textbf{Accuracy} & \textbf{Precision} & \textbf{Recall} & \textbf{F1-score} & \textbf{AUC-ROC} & \textbf{Time}\\%
        \hline
        {DNN} & 0.9870 & \textbf{0.9974} & 0.8425 & 0.9135 & 0.9687 & 0.0577\\%
        \hline
        {Wide and Deep Network} & 0.9867 & 0.9955 & 0.8393 & 0.9108 & 0.9643 & 0.0612\\%
        \hline
        {LSTM} & 0.9891 & 0.9905 & 0.8744 & 0.9288 & 0.9791 & 0.1533\\%
        \hline
        {Attention-Based Network} & 0.9816 & 0.9667 & 0.8003 & 0.8757 & 0.9378 & 0.1119\\%
        \hline
        {Cross Network} & 0.9870 & 0.9942 & 0.8451 & 0.9136 & 0.9708 & \textbf{0.0576}\\%
        \hline
        {FM} & 0.9733 & 0.8975 & 0.7581 & 0.8219 & 0.9113 & 0.0592\\%
        \hline
        {Current Study} & \textbf{0.9952} & 0.9933 & \textbf{0.9484} & \textbf{0.9703} & \textbf{0.9987} & \textbf{0.0576}\\%
        \hline
    \end{tabular}%
    }%
\end{table}%

\subsection{Comparative Analysis With Other State-of-the-Art Models}%
\label{CAWOSAM}%
The proposed framework is evaluated against several state-of-the-art models to further support its efficiency. Table~\ref{resultf} shows that various models trained on the dataset introduced in \cite{Kang2021} have been compared, where all baseline approaches yield lower accuracy and F1-scores compared with the proposed framework. An exception is observed in \cite{Alalwany2022}, where SMOTE was applied unsystematically to the entire dataset, including the test set. This evaluation protocol is methodologically inappropriate, as synthetic oversampling should be restricted to the training set to maintain the natural distribution of the test data for generalization. In contrast, the evaluation of the proposed framework is conducted on an imbalanced test set, preserving the real-world data distribution. Table~\ref{resultf2} presents a comparative analysis using the dataset in \cite{Song2020}, where the proposed framework achieves superior performance across all attack categories. Notably, in the fuzzy attack scenario, the competing methods exhibit substantial performance degradation, while the proposed framework consistently achieves optimal classification metrics.%

\begin{table}[t]
    \centering
    \caption{Performance Comparison for Overall Framework Trained on \cite{Kang2021}}%
    \label{resultf}%
    \resizebox{\columnwidth}{!}{%
    \begin{tabular}{|l|l|l|l|l|l|}
        \hline
        \textbf{Model} & \textbf{Accuracy} & \textbf{Precision} & \textbf{Recall} & \textbf{F1-score}\\%
        \hline
        Stacking \cite{Alalwany2022} & 0.9847 & 0.9804 & \textbf{0.9891} & \textbf{0.9847}\\%
        \hline
        DT \cite{Aksu2022} & 0.9895 & 0.9854 & 0.8852 & 0.9326\\%
        \hline
        XGBoost \cite{Raj2023} & 0.9450 & 0.9650 & 0.9250 & 0.9450\\%
        \hline
        KANs \cite{Liu2024} & 0.9490 & 0.9921 & 0.5191 & 0.6816\\%
        \hline
        Current Study & \textbf{0.9952} & \textbf{0.9933} & 0.9484 & 0.9703\\%
        \hline
    \end{tabular}%
    }%
\end{table}%

\begin{table}[t]
    \centering
    \caption{Performance Comparison for Overall Framework Trained on \cite{Song2020}}%
    \label{resultf2}%
    \resizebox{\columnwidth}{!}{%
    \begin{tabular}{|l|l|l|l|l|l|}
        \hline
        \textbf{Attack Type} & \textbf{Model} & \textbf{Accuracy} & \textbf{Precision} & \textbf{Recall} & \textbf{F1-score}\\%
        \hline
        \multirow{4}{*}{DoS} & Generator-Detector \cite{Song2021} & 0.9869 & 0.9751 & 0.9988 & 0.9833\\%
                             & Reduced Inception-ResNet \cite{Song2020} & 0.9997 & 1.0000 & 0.9989 & 0.9995\\%
                             & ConvLSTM-GNB \cite{Mansourian2023}& 1.0000 & 1.0000 & 1.0000 & 1.0000\\%
                             & Current Study & 1.0000 & 1.0000 & 1.0000 & 1.0000\\%
        \hline
        \multirow{4}{*}{Fuzzy} & Generator-Detector \cite{Song2021} & 0.9387 & 0.9445 & 0.9626 & 0.9305\\%
                               & Reduced Inception-ResNet \cite{Song2020} & 0.9982 & 0.9995 & 0.9965 & 0.9980\\%
                               & ConvLSTM-GNB \cite{Mansourian2023}& 1.0000 & 0.9960 & 0.9970 & 0.9970\\%
                               & Current Study & 1.0000 & 1.0000 & 1.0000 & 1.0000\\%
        \hline
        \multirow{4}{*}{Gear Spoofing} & Generator-Detector \cite{Song2021} & 0.9306 & 0.9768 & 0.8803 & 0.9261\\%
                                       & Reduced Inception-ResNet \cite{Song2020} & 0.9995 & 0.9999 & 0.9989 & 0.9994\\%
                                       & ConvLSTM-GNB \cite{Mansourian2023} & 1.0000 & 1.0000 & 1.0000 & 1.0000\\%
                                       & Current Study & 1.0000 & 1.0000 & 1.0000 & 1.0000\\%
        \hline
        \multirow{4}{*}{RPM Spoofing} & Generator-Detector \cite{Song2021} & 0.9997 & 1.0000 & 0.9997 & 0.9992\\%
                                      & Reduced Inception-ResNet \cite{Song2020} & 0.9997 & 0.9999 & 0.9994 & 0.9996\\%
                                      & ConvLSTM-GNB \cite{Mansourian2023} & 1.0000 & 1.0000 & 1.0000 & 1.0000\\%
                                      & Current Study & 1.0000 & 1.0000 & 1.0000 & 1.0000\\%
        \hline
    \end{tabular}%
    }%
\end{table}%

\subsection{Discussion}%
The most significant strength of the proposed framework is its remarkably high accuracy and F1-score compared with other models, which has been statistically validated in terms of reliability and robustness. Additionally, the proposed framework demonstrates superior performance in inference time and ROC-AUC. The principal limitation of the proposed framework lies in its reliance on DT to identify the optimal subspace during the execution of the 2P-GA. This limitation primarily arises from hardware constraints. DT fell short in fully leveraging the capacity of spatial features, ultimately selecting only two of them. It is believed that with access to more advanced hardware and the use of more sophisticated classifiers, such as RF, improved performance could be achieved.

\section{Conclusion and Future Works}%
\label{sec:s05-conclusion-and-future-works}%
Controller Area Network (CAN) bus remains a critical vulnerability in intelligent transportation systems (ITS) due to its lack of inherent security mechanism. While the existing anomaly detection solutions address isolated attack patterns, they often neglect the key structures of attacks, leading to a lack of reliability and robustness. To address these shortcomings, this study proposes a spatiotemporal fusion strategy framework that integrates spatial and temporal information using cascade feature-level fusion optimized via a two-parameter genetic algorithm (2P-GA). To evaluate the effectiveness of the proposed framework, extensive experiments were conducted, and a paired \(t\)-test was performed, resulting in $t = 22.83$, $p = 1.17 \times 10^{-6}$, which confirms that the observed performance improvement is statistically significant. An accuracy of \(0.9952\), a precision of \(0.9933\), a recall of \(0.9484\), an F1-score of \(0.9703\), and an AUC-ROC of \(0.9987\) were achieved, demonstrating the superiority of the proposed method over baseline approaches. The proposed model was able to achieve \(100\%\) accuracy across all attack types in the CAR-HACKING dataset, thereby establishing itself as the most effective solution reported so far for mitigating security threats in the CAN bus of modern vehicles.%

Potential directions for future research related to this study include the following:%
\begin{itemize}
    \item \textbf{Addressing the limitations of the decision tree (DT) used in the genetic algorithm for subspace selection.} By replacing DT with a more powerful classifier, such as random forest (RF), the model could better leverage spatial and temporal features, thus improving feature selection and overall performance. This approach, supported by advanced hardware, would enhance the accuracy and robustness of the model in detecting CAN bus attacks.%
    \item \textbf{Introducing a Spatial Data Prediction Model that utilizes Deep Q-Learning to generate larger errors when encountering attacks and smaller errors for normal messages.} This enhancement can improve the cascade feature-level fusion strategy, enabling it to operate with better performance.%
\end{itemize}

\bibliographystyle{IEEEtran}%
\bibliography{references}%

\vspace{-33pt}


\begin{IEEEbiographynophoto}{Mohammad {Fatahi}}
received his M.Sc. degree in Control Engineering in 2023 from the University of Tehran, Iran. His research interests include machine learning, computer vision, control theory, and signal processing.%
\end{IEEEbiographynophoto}%

\vspace{-33pt}

\begin{IEEEbiographynophoto}{Danial {Sadrian Zadeh}}
(Graduate Student Member, IEEE) received his B.Sc. degree in Control Engineering in 2018 from the Petroleum University of Technology, Iran, and his M.Sc. degree in Control Engineering in 2021 from the University of Tehran, Iran. He is now pursuing his Ph.D. in the area of Pattern Analysis and Machine Intelligence in the Department of Electrical and Computer Engineering at the University of Waterloo, Canada. His research interests include autonomous vehicles, control theory, machine learning, sensor data fusion, and state estimation.%
\end{IEEEbiographynophoto}%

\vspace{-33pt}

\begin{IEEEbiographynophoto}{Benyamin {Ghojogh}}
received his B.Sc. degree in Electrical Engineering from the Amirkabir University of Technology, Tehran, Iran, in 2015, his M.Sc. degree in Electrical Engineering from the Sharif University of Technology, Tehran, Iran, in 2017, and his Ph.D. in Electrical and Computer Engineering (in the area of Pattern Analysis and Machine Intelligence) from the University of Waterloo, Waterloo, ON, Canada, in 2021. He was a postdoctoral fellow, focusing on machine learning, at the University of Waterloo in 2021. His research interests include machine learning, dimensionality reduction, manifold learning, computer vision, data science, and deep learning.%
\end{IEEEbiographynophoto}%

\vspace{-33pt}

\begin{IEEEbiographynophoto}{Behzad {Moshiri}}
(Senior Member, IEEE) received his M.Sc. and Ph.D. degrees in control systems engineering from the University of Manchester Institute of Science and Technology (UMIST) in 1987 and 1991, respectively. He is currently a full professor of control systems engineering at the School of Electrical and Computer Engineering, University of Tehran, Iran. He has also been an adjunct professor in the Department of Electrical and Computer Engineering at the University of Waterloo, Canada, since 2014. He is now serving as the chairperson of the IEEE Control Systems Chapter in the IEEE Iran Section since December 2018. He has been a member of the International Society of Information Fusion (ISIF) since 2002, a senior member of IEEE since 2006, and a member of Waterloo AI Institute since 2019. He is the author/co-author of more than 360 articles, including 130+ journal papers and 20+ book chapters. His fields of research include advanced industrial control, advanced instrumentation systems, data fusion theory, and its applications in areas such as robotics, process control, mechatronics, information technology (IT), intelligent transportation systems (ITS), bioinformatics, and financial engineering.%
\end{IEEEbiographynophoto}%

\vspace{-33pt}

\begin{IEEEbiographynophoto}{Otman {Basir}}
(Member, IEEE) is the Associate Director of the University of Waterloo's Research Center on Pattern Recognition and Machine Intelligence. His research program is concerned with the design of Self-Organizing and Operating Systems, Sensor Fusion, and Human-Machine Engagement Systems.%
\end{IEEEbiographynophoto}%

\vfill

\end{document}